\documentstyle[a4wide,11pt]{article}

\def\xp{%
 \par\hangafter=1\hangindent=2em%
 \parskip 0.2ex plus 0.1ex minus 0.1ex%
 \noindent}

\newcommand{\lhiparrow}{%
 \ifmmode{{\sim\sim\!>}}%
 \else{${\sim\sim\!>}$}%
 \fi}

\title{LHIP: Extended DCGs for Configurable Robust Parsing\thanks{%
	This work was carried out under grants nos.\ 20-33903.92 and
	12-36505.92 from the Swiss National Fund.}}

\author{Afzal Ballim\hspace{3em}Graham Russell\\[0.5ex]
        ISSCO, University of Geneva\\
	54 Route des Acacias, Geneva, Switzerland\\
	\small email: \tt afzal@divsun.unige.ch, russell@divsun.unige.ch}

\begin{document}

\maketitle

\begin{abstract}
\noindent
We present LHIP, a system for incremental grammar development using an
extended DCG formalism.  The system uses a robust island-based parsing
method controlled by user-defined performance thresholds.\\[0.5ex]
\end{abstract}

\section{LHIP Overview}

This paper describes LHIP (Left-Head corner Island Parser), a parser designed
for broad-coverage handling of unrestricted text.  The system interprets an
extended DCG formalism to produce a robust analyser that finds parses of the
input made from `islands' of terminals (corresponding to terminals consumed
by successful grammar rules).  It is currently in use for processing dialogue
transcripts from the HCRC Map Task Corpus (Anderson et al., 1991), although we
expect its eventual applications to be much wider.%
\footnote{
Note that the input consists of {\em written\/} texts within the Map Task
Corpus; LHIP is not intended for use in speech processing.
}
Transcribed natural speech contains a number of frequent characteristic
`ungrammatical' phenomena:  filled pauses, repetitions, restarts, etc.\ (as in
e.g.\ {\em Right I'll have \ldots you know, like I'll have to \ldots so I'm
going between the picket fence and the mill, right.\/}).%
\footnote{
This example is taken from the Map Task Corpus.
}
While a full analysis of a conversation might well take these into account,
for many purposes they represent a significant obstacle to analysis.  LHIP
provides a processing method which allows selected portions of the input to be
ignored or handled differently.

The chief modifications to the standard Prolog `grammar rule' format are of
two types:  one or more right-hand side (RHS) items may be marked as `heads',
and one or more RHS items may be marked as `ignorable'.  We expand on these
points and introduce other differences below.

The behaviour of LHIP can best be understood in terms of the
notions of {\bf island}, {\bf span}, {\bf cover} and {\bf threshold}:
\begin{description}

\item[Island:]
Within an input string consisting of the terminals
$\langle t_1, t_2, \ldots t_n\rangle$, an island is a subsequence
$\langle t_i, t_{i+1}, \ldots t_{i+m}\rangle$, whose {\bf length} is $m + 1$.

\item[Span:]

The span of a grammar rule $R$ is the length of the longest island
$\langle t_i, \ldots t_j\rangle$ such that terminals $t_i$ and $t_j$ are
both consumed (directly or indirectly) by $R$.

\item[Cover:]
A rule $R$ is said to cover $m$ items if $m$ terminals are consumed within
the island described by $R$.  The {\bf coverage} of $R$ is then $m$.

\item[Threshold:]
The threshold of a rule is the minimum value for the ratio of its
coverage $c$ to its span $s$ which must hold in order for the rule to
succeed.
Note that $c \leq s$, and that if $c = s$ the rule has completely covered
the span, consuming all terminals.

\end{description}
As implied here, rules need not cover all of the input in order to succeed.
More specifically, the constraints applied in creating islands are such that
islands do not have to be adjacent, but may be separated by non-covered input.
Moreover, an island may itself contain input which is unaccounted for by the
grammar.  Islands do not overlap, although when multiple analyses exist they
will in general involve different segmentations of the input into islands.

There are two notions of non-coverage of the input:  {\bf sanctioned} and {\bf
unsanctioned} non-coverage.  The latter case arises when the grammar simply
does not account for some terminal.  Sanctioned non-coverage means that
some number of special `ignore' rules have been applied which simulate
coverage of input material lying between the islands, thus in effect making
the islands contiguous.  Those parts of the input that have been `ignored'
are considered to have been consumed.  These ignore rules can be invoked
individually or as a class.  It is this latter capability which distinguishes
ignore rules from regular rules, as they are functionally equivalent
otherwise, mainly serving as a notational aid for the grammar writer.

Strict adjacency between RHS clauses can be specified in the grammar.  It is
possible to define global and local thresholds for the proportion of the
spanned input that must be covered by rules; in this way, the user of an
LHIP grammar can exercise quite fine control over the required accuracy and
completeness of the analysis.

A chart is kept of successes and failures of rules, both to improve efficiency
and to provide a means of identifying unattached constituents.  In addition,
feedback is given to the grammar writer on the degree to which the grammar is
able to cope with the given input; in a context of grammar development, this
may serve as notification of areas to which the coverage of the grammar might
next be extended.

The notion of `head' employed here is connected more closely with processing
control than linguistics.  In particular, nothing requires that a head of a
rule should share any information with the LHS item, although in practice it
often will.  Heads serve as anchor-points in the input string around which
islands may be formed, and are accordingly treated before non-head items
(RHS items are re-ordered during compilation\,--\,see below).  In the
central role of heads, LHIP resembles parsers devised by Kay (1989) and
van~Noord (1991); in other respects, including the use which is made of heads,
the approaches are rather different, however.

\section{The LHIP System}

In this section we describe the LHIP system.  First, we define what
constitutes an acceptable LHIP grammar, second, we describe the process of
converting such a grammar into Prolog code, and third, we describe the
analysis of input with such a grammar.

LHIP grammars are an extended form of Prolog DCG grammars.  The extensions
can be summarized as follows:%
\footnote{
A version of LHIP exists which permits a form of negation on RHS clauses.
That version is not described here.
}
\begin{enumerate}
\item
one or more RHS clauses may be nominated as heads;
\item
one or more RHS clauses may be marked as optional;
\item
`ignore' rules may be invoked;
\item
adjacency constraints may be imposed between RHS clauses;
\item
a global threshold level may be set to determine the minimum fraction of
spanned input that may be covered in a parse, and
\item
a local threshold level may be set in a rule to override the global threshold
within that rule.
\end{enumerate}
We provide a syntactic definition (below) of a LHIP grammar rule, using a
notation with syntactic rules of the form $C \Rightarrow F_1 \mid F_2 \ldots
\mid F_n$ which indicates that the category $C$ may take any of the forms
$F_1$ to $F_n$.  An optional item in a form is denoted by surrounding it
with square brackets `[\,\ldots]'.  Syntactic categories are {\em italicised},
while terminals are underlined: `$\underline{\ldots}$'.

A LHIP grammar rule has the form:

\smallskip
\begin{tabbing}
\makebox[0.5em]{}{\it lhiprule} $\Rightarrow$
[ $\underline{\bf -}$ ] {\it term} [ $\underline{\#}\; T$ ]
$\underline{\lhiparrow}$  {\it lhipbody}
\end{tabbing}
\smallskip

\noindent
where $T$ is a value between zero and one.  If present, this value defines the
local threshold fraction for that rule.

This local threshold value overrules the global threshold.
The symbol `${\bf -}$' before the name of a rule marks it as being an
`ignore' rule.
Only a rule defined this way can be invoked as an ignore rule in an RHS
clause.

\smallskip
\begin{tabbing}
\makebox[1.5em]{}\={\it lhipbody} $\Rightarrow$ \= {\it lhipclause} \\
\> \> $\mid \; {\it lhipclause} \; \underline{\bf ,} \; {\it lhipbody}$\\
\> \> $\mid \; {\it lhipclause} \; \underline{\bf ;} \; {\it lhipbody}$\\
\> \> $\mid \; {\it lhipclause} \; \underline{\bf :} \; {\it lhipbody}$\\
\> \> $\mid \; \underline{\bf (?} \; {\it lhipbody} \; \underline{\bf ?)}$
\end{tabbing}
\smallskip

The connectives `{\bf ,}' and `{\bf ;}' have the same precedence as in Prolog,
while `{\bf :}' has the same precedence as `{\bf ,}'.  Parentheses may be
used to resolve ambiguities.
The connective `{\bf ,}' is used to indicate that strings subsumed by two
RHS clauses are ordered but not necessarily adjacent in the input.  Thus
`{\em A {\bf,} B\/}' indicates that {\em A} precedes {\em B} in the input,
perhaps with some intervening material.  The stronger constraint of
immediate precedence is marked by `{\bf :}'; `{\em A {\bf :} B\/}'
indicates that the span of $A$ precedes that of $B$, and that there is no
uncovered input between the two.
Disjunction is expressed by `{\bf ;}', and optional RHS clauses are
surrounded by `{\bf (?{\rm\,\ldots}?)}'.

\begin{tabbing}
\makebox[1.5em]{}\={\it lhipclause} $\Rightarrow$ \= {\it term} \\
\> \> $\mid \; \underline{\bf *} \; {\it term}$ \\
\> \> $\mid \; \underline{\bf @} \; {\it string}$\\
\> \> $\mid \; \underline{\bf * @} \; {\it string}$\\
\> \> $\mid \; \underline{\bf -} \; {\it term}$\\
\> \> $\mid \; \underline{[\:]}$\\
\> \> $\mid \; \underline{\{} \; {\it prologcode} \; \underline{\}}$
\end{tabbing}

\smallskip

The symbol `{\bf *}' is used to indicate a head clause.  A rule name is a
Prolog $term$, and only rules and terminal items may act as heads within a
rule body.  The symbol `{\bf @}' introduces a terminal $string$.  As
previously said, the purpose of ignore rules is simply to consume input
terminals, and their intended use is in facilitating repairs in analysing
input that contains the false starts, restarts, filled pauses, etc.\ mentioned
above.  These rules are referred to individually by preceding their name by
the `${\bf -}$' symbol.  They can also be referred to as a class in a rule
body by the special RHS clause `$[\:]$'.  If used in a rule body, they
indicate that input is {\em potentially} ignored\,--\,the problems that ignore
rules are intended to repair will not always occur, in which case the rules
succeed without consuming any input.  There is a semantic restriction on the
body of a rule which is that it must contain at least one clause which
necessarily covers input (optional clauses and ignore rules do not necessarily
cover input).

The following is an example of a LHIP rule.  Here, the sub-rule
`{\sf conjunction(Conj)}' is marked as a head and is therefore evaluated
before either of `{\sf s(Sl)}' or `{\sf s(Sr)}':

\smallskip
\begin{tabbing}
\makebox[6em]{}\=\kill
\sf\makebox[3.5em]{}s(conjunct(Conj, Sl, Sr)) \lhiparrow \\
\>\sf  s(Sl), \\
\>\sf  * conjunction(Conj),\\
\>\sf  s(Sr).
\end{tabbing}
\smallskip

How is such a rule converted into Prolog code by the LHIP system?  First, the
rule is read and the RHS clauses are partitioned into those marked as heads,
and those not.  A record is kept of their original ordering, and this record
allows each clause to be constrained with respect to the clause that precedes
it, as well as with respect to the next head clause which follows it.
Additional code is added to maintain a chart of known successes and failures
of each rule.  Each rule name is turned into the name of a Prolog clause, and
additional arguments are added to it.  These arguments are used for the input,
the start and end points of the area of the input in which the rule may
succeed, the start and end points of the actual part of the input over which
it in fact succeeds, the number of terminal items covered within that island,
a reference to the point in the chart where the result is stored, and a list
of pointers to sub-results.  The converted form of the above rule is given
below (minus the code for chart maintenance):

\smallskip
\begin{verbatim}
  s(conjunct(H,I,J), A, B, C, D, E, F,
    [L|K]-K, G) :-
      lhip_threshold_value(M),
      conjunction(H, A, B, C, O, P, Q,
                  R-S, _),
      s(I, A, B, O, D, _, T, G-R, _),
      s(J, A, P, C, _, E, U, S-[], _),
      F is U+Q+T,
      F/(E-D)>=M.
\end{verbatim}
\smallskip

The important points to note about this converted form are the following:
\begin{enumerate}

\item
the {\tt conjunction} clause is searched for before either of the two {\tt s}
clauses;

\item
the region of the input to be searched for the {\tt conjunction} clause is the
same as that for the rule's LHS ({\tt B}--{\tt C}): its island extends from
{\tt O} to {\tt P} and covers {\tt Q} items;

\item
the search region for the first {\tt s} clause is {\tt B}--{\tt O} (i.e.\ from
the start of the LHS search region to the start of the {\tt conjunction}
island), its island starts at {\tt D} and covers {\tt T} items;

\item
the search region for the second {\tt s} clause is {\tt P}--{\tt C}
(i.e.\ from the end of the {\tt conjunction} island to the end of the LHS
search region), its island ends at {\tt E} and covers {\tt U} items;

\item
the island associated with the rule as a whole extends from {\tt D} to
{\tt E} and covers {\tt F} items, where {\tt F} is {\tt U} + {\tt Q} + {\tt
T};

\item
\verb|lhip_threshold_value/1| unifies its argument {\tt M} with the current
global threshold value.

\end{enumerate}
In the current implementation of LHIP, compiled rules are interpreted
depth-first and left-to-right by the standard Prolog theorem-prover, giving an
analyser that proceeds in a top-down, `left-head-corner' fashion.  Because of
the reordering carried out during compilation, the situation regarding
left-recursion is slightly more subtle than in a conventional DCG\@.  The `{\sf
s(conjunct(\ldots))}' rule shown above is a case in point.  While at first
sight it appears left-recursive, inspection of its converted form shows its
true leftmost subrule to be `{\sf conjunction}'.  Naturally, compilation may
induce left-recursion as well as eliminating it, in which case LHIP will
suffer from the same termination problems as an ordinary DCG formalism
interpreted in this way.  And as with an ordinary DCG formalism, it is
possible to apply different parsing methods to LHIP in order to circumvent
these problems (see e.g.\ Pereira and Shieber, 1987).  A related issue
concerns the interpretation of embedded Prolog code.  Reordering of RHS
clauses will result in code which precedes a head within a LHIP rule being
evaluated after it; judicious freezing of goals and avoidance of unsafe cuts
are therefore required.

LHIP provides a number of ways of applying a grammar to input.  The simplest
allows one to enumerate the possible analyses of the input with the grammar.
The order in which the results are produced will reflect the lexical
ordering of the rules as they are converted by LHIP\@.  With the threshold
level set to 0, all analyses possible with the grammar by deletion of input
terminals can be generated.  Thus, supposing a suitable grammar, for the
sentence {\em John saw Mary and Mark saw them\/} there would be analyses
corresponding to the sentence itself, as well as {\em John saw Mary},
{\em John saw Mark}, {\em John saw them}, {\em Mary saw them},
{\em Mary and Mark saw them,} etc.

By setting the threshold to 1, only those partial analyses that have no
unaccounted for terminals within their spans can succeed.  Hence, {\em Mark
saw them\/} would receive a valid analysis, as would {\em Mary and Mark saw
them\/}, provided that the grammar contains a rule for conjoined NPs; {\em
John saw them\/}, on the other hand, would not.  As this example
illustrates, a partial analysis of this kind may not in fact correspond to a
true sub-parse of the input (since {\em Mary and Mark\/} was not a conjoined
subject in the original).  Some care must therefore be taken in interpreting
results.

A number of built-in predicates are provided which allow the user to constrain
the behaviour of the parser in various ways, based on the notions of coverage,
span and threshold:

\smallskip

\xp
\verb|lhip_phrase(+C,+S)|\\
Succeeds if the input {\tt S} can be parsed as an instance of category
{\tt C}.

\xp
\verb|lhip_cv_phrase(+C,+S)|\\
As for \verb|lhip_phrase/2|, except that all of the input must be covered.

\xp
\verb|lhip_phrase(+C,+S,-B,-E,-Cov)|\\
As for \verb|lhip_phrase/2|, except that {\tt B} binds to the
beginning of the island described by this application of {\tt C}, {\tt
E} binds to the position immediately following the end, and {\tt Cov} binds
to the number of terminals covered.

\xp
\verb|lhip_mc_phrases(+C,+S,-Cov,-Ps)|\\
The maximal coverage of {\tt S} by {\tt C} is {\tt Cov}.  {\tt Ps} is
the set of parses of {\tt S} by {\tt C} with coverage {\tt Cov}.

\xp
\verb|lhip_minmax_phrases(+C,+S,-Cov,-Ps)|\\
As for \verb|lhip_mc_phrases/4|, except that {\tt Ps} is additionally
the set of parses with the least span.

\xp
\verb|lhip_seq_phrase(+C,+S,-Seq)|\\
Succeeds if {\tt Seq} is a sequence of one or more parses of {\tt S} by
{\tt C} such that they are non-overlapping and each consumes input
that precedes that consumed by the next.

\xp
\verb|lhip_maxT_phrases(+C,+S,-MaxT)|\\
{\tt MaxT} is the set of parses of {\tt S} by {\tt C} that have the highest
threshold value. On backtracking it returns the set with the next highest
threshold value.

\smallskip

\noindent
In addition, other predicates can be used to search the chart for constituents
that have been identified but have not been attached to the parse tree.
These include:

\smallskip

\xp
\verb|lhip_success|\\
Lists successful rules, indicating island position and coverage.

\xp
\verb|lhip_ms_success|\\
As for \verb|lhip_success|, but lists only the most specific successful
rules (i.e.\ those which have themselves succeeded but whose results have
not been used elsewhere).

\xp
\verb|lhip_ms_success(N)|\\
As for \verb|lhip_ms_success|, but lists only successful instances of rule
{\tt N}.

\smallskip

\noindent
Even if a sentence receives no complete analysis, it is likely to contain some
parsable substrings; results from these are recorded together with their
position within the input.  By using these predicates, partial but possibly
useful information can be extracted from a sentence despite a global failure
to parse it (see section~\ref{partial-results}).

The conversion of the grammar into Prolog code means that the user of the
system can easily develop analysis tools that apply different constraints,
using the tools provided as building blocks.

\section{Using LHIP}

As previously mentioned, LHIP facilitates a cyc\-lic approach to grammar
development.  Suppose one is writing an English grammar for the Map Task
Corpus, and that the following is the first attempt at a rule for noun
phrases (with appropriate rules for determiners and nouns):

\smallskip

\begin{tabbing}
\makebox[6em]{}\=\kill
\sf\makebox[3.5em]{}np(N, D, A) \# 0.5 $\lhiparrow$ \\
\>\sf determiner(D), \\
\>\sf * noun(N).
\end{tabbing}

\smallskip

While this rule will adequately analyse simple NPs such as {\em your map},
or {\em a missionary camp}, on a NP such as {\em the bottom right-hand
corner\/} it will give analyses for {\em the bottom}, {\em the right-hand\/}
and {\em the corner}.  Worse still, in a long sentence it will join
determiners from the start of the sentence to nouns that occur in the latter
half of the sentence.  The number of superfluous analyses can be reduced by
imposing a local threshold level, of say 0.5.  By looking at the various
analyses of sentences in the corpus, one can see that this rule gives the
skeleton for noun phrases, but from the fraction of coverage of these parses
one can also see that it leaves out an important feature, adjectives, which
are optionally found in noun phrases.

\smallskip

\begin{samepage}
\begin{tabbing}
\makebox[6em]{}\=\kill
\sf\makebox[3.5em]{}np(N, D, A) \# 0.5 $\lhiparrow$ \\
\>\sf determiner(D), \\
\>\sf (? adjectives(A) ?),\\
\>\sf * noun(N).
\end{tabbing}
\end{samepage}

\smallskip

With this rule, one can now handle such phrases as {\em the left-hand bottom
corner}, and {\em a banana tree}.  Suppose further that this rule is now
applied to the corpus, and then the rule is applied again but with a local
threshold level of 1.  By looking at items parsed in the first case but not
in the second, one can identify features of noun phrases found in the corpus
that are not covered by the current rules.  This might include, for
instance, phrases of the form {\em a slightly dipping line}.  One can then go
back to the grammar and see that the noun phrase rule needs to be changed to
account for certain types of modifier including adjectives and adverbial
modifiers.

It is also possible to set local thresholds dynamically, by making use of
the `\{\,prolog code\,\}' facility:

\smallskip

\begin{tabbing}
\makebox[6em]{}\=\kill
\sf\makebox[3.5em]{}np(N, D, A) \# T $\lhiparrow$ \\
\>\sf determiner(D), \\
\>\sf (? adjectives(A) ?),\\
\>\sf * noun(N),\\
\>\sf \{\,set\_dynamic\_threshold(A,T)\,\}.
\end{tabbing}

\smallskip

\noindent
In this way, the strictness of a rule may be varied according to information
originating either within the particular run-time invocation of the rule, or
elsewhere in the current parse.  For example, it would be possible, by
providing a suitable definition for {\sf set\_dynamic\_threshold/2}, to set
{\sf T} to 0.5 when more than one optional adjective has been found, and 0.9
otherwise.

Once a given rule or set of rules is stable, and the writer is satisfied with
the performance of that part of the grammar, a local threshold value of 1 may
be assigned so that superfluous parses will not interfere with work elsewhere.

The use of the chart to store known results and failures allows the user to
develop hybrid parsing techniques, rather than relying on the default
depth-first top-down strategy given by analysing with respect to the
top-most category.  For instance, it is possible to analyse the input in
`layers' of linguistic categories, perhaps starting by analysing
noun-phrases, then prepositions, verbs, relative clauses, clauses,
conjuncts, and finally complete sentences.  Such a strategy allows the user
to perform processing of results between these layers, which can be useful
in trying to find the `best' analyses first.

\section{Partial results}\label{partial-results}

The discussion of built-in predicates mentioned facilities for recovering
partial parses.  Here we illustrate this process, and indicate what further
use might be made of the information thus obtained.

In the following example, the chart is inspected to reveal what constituents
have been built during a failed parse of the truncated sentence {\em Have you
the tree by the brook that\,\ldots}:

\smallskip

{\small\begin{verbatim}
    > lhip_phrase(s(S),
        [have,you,the,tree,by,the,brook,that]).
    no
    > lhip_success.
    (-1) [7--8) /1 ~~> @brook
    (-1) [5--6) /1 ~~> @by
    (-1) [1--2) /1 ~~> @have
    (-1) [8--9) /1 ~~> @that
    (-1) [3--4) /1 ~~> @the
    (-1) [6--7) /1 ~~> @the
    (-1) [4--5) /1 ~~> @tree
    (-1) [2--3) /1 ~~> @you
    (4) [2--8) /4 ~~> np(nppp(you,pp(by,np(the,brook,B))))
    (4) [3--8) /5 ~~> np(nppp(np(the,tree,C),pp(by,np(the,brook,D))))
    (5) [3--8) /2 ~~> np(np(the,brook,A))
    (5) [6--8) /2 ~~> np(np(the,brook,G))
    (5) [3--5) /2 ~~> np(np(the,tree,E))
    (7) [4--5) /1 ~~> noun(tree)
    (8) [7--8) /1 ~~> noun(brook)
    (9) [2--3) /1 ~~> np(you)
    (10) [5--8) /3 ~~> pp(pp(by,np(the,brook,F)))
    (11) [3--4) /1 ~~> det(the)
    (11) [6--7) /1 ~~> det(the)
    yes
\end{verbatim}}

\smallskip

\noindent
Each rule is listed with its identifier (`{\tt -1}' for lexical rules), the
island which it has analysed with beginning and ending positions, its
coverage, and the representation that was constructed for it.  From this
output it can be seen that the grammar manages reasonably well with noun
phrases, but is unable to deal with questions (the initial auxiliary {\em
have} remains unattached).

Users will often be more interested in the successful application of rules
which represent maximal constituents.  These are displayed by
\verb|lhip_ms_success|:

\smallskip

{\small\begin{verbatim}
    > lhip_ms_success.
    (-1) [1--2) /1 ~~> @have
    (-1) [8--9) /1 ~~> @that
    (4) [2--8) /4 ~~> np(nppp(you,pp(by,np(the,brook,J))))
    (4) [3--8) /5 ~~> np(nppp(np(the,tree,H),pp(by,np(the,brook,I))))
    (5) [3--8) /2 ~~> np(np(the,brook,K))
    yes
\end{verbatim}}

\smallskip

\noindent
Here, two unattached lexical items have been identified, together with two
instances of rule 4, which combines a NP with a postmodifying PP\@.  The first
of these has analysed the island {\em you the tree by the brook}, ignoring
{\em the tree}, while the second has analysed {\em the tree by the brook},
consuming all terminals.  There is a second analysis for {\em the tree by the
brook}, due to rule 5, which has been obtained by ignoring the sequence {\em
tree by the}.  From this information, a user might wish to rank the three
results according to their respective span:coverage ratios, probably preferring
the second.

\section{Discussion}

The ability to deal with large amounts of possibly ill-formed text is one of
the principal objectives of current NLP research.  Recent proposals include
the use of probabilistic methods (see e.g.\ Briscoe and Carroll, 1993) and
large robust deterministic systems like Hindle's Fidditch (Hindle, 1989).%
\footnote{
Indeed, the ability of Fidditch to return a sequence of parsed but unattached
phrases when a global analysis fails has clearly influenced the design of LHIP.
}
Experience so far suggests that systems like LHIP may in the right
circumstances provide an alternative to these approaches.  It combines the
advantages of Prolog-interpreted DCGs (ease of modification, parser output
suitable for direct use by other programs, etc.)\ with the ability to relax
the adjacency constraints of that formalism in a flexible and dynamic manner.

LHIP is based on the assumption that partial results can be useful (often much
more useful than no result at all), and that an approximation to complete
coverage is more useful when it comes with indications of how approximate it
is.  This latter point is especially important in cases where a grammar must
be usable to some degree at a relatively early stage in its development, as
is, for example, the case with the development of a grammar for the Map Task
Corpus.  In the near future, we expect to apply LHIP to a different problem,
that of defining a restricted language for specialized parsing.

The rationale for the distinction between sanctioned and unsanctioned
non-coverage of input is twofold.  First, the `ignore' facility
permits different categories of unidentified input to be distinguished.  For
example, it may be interesting to separate material which occurs at the
start of the input from that appearing elsewhere.  Ignore rules have a
similar functionality to that of normal rules.  In particular, they can have
arguments, and may therefore be used to assign a structure to unidentified
input so that it may be flagged as such within an overall parse.  Secondly,
by setting a threshold value of 1, LHIP can be made to perform like a
standardly interpreted Prolog DCG, though somewhat more efficiently due to
the use of the chart.%
\footnote{
In large grammars there is a significant time gain.  The chart's main
advantage, however, is in identifying unattached constituents and allowing
a `layered' approach to analysis of input.
}

A number of possible extensions to the system can be envisaged.  Whereas at
present each rule is compiled individually, it would be preferable to enhance
preprocessing in order to compute certain kinds of global information from the
grammar.  One improvement would be to determine possible linking of
`root-to-head' sequences of rules, and index these to terminal items for use
as an oracle during analysis.  A second would be to identify those items whose
early analysis would most strongly reduce the search space for subsequent
processing and scan the input to begin parsing at those points rather than
proceeding strictly from left to right.  This further suggests the possibility
of a parallel approach to parsing.  We expect that these measures would
increase the efficiency of LHIP\@.

Currently, also, results are returned in an order determined by the order of
rules in the grammar.  It would be preferable to arrange matters in a more
cooperative fashion so that the best (those with the highest coverage to span
ratio) are displayed first.  Support for bidirectional parsing (see Satta
and Stock, to appear) is another candidate for inclusion in a later version.
These appear to be longer-term research goals, however.\footnote{
Source code for the LHIP system has been made publicly available. For
information, contact the authors.}

\paragraph{Acknowledgments:}
The authors would like to thank Louis des
Tombe and Dominique Estival for comments on earlier versions of this paper.

\section*{References}

\xp
Anderson, A.H., M. Bader, E.G. Bard, E. Boyle, G. Doherty, S. Garrod, S.
Isard, J. Kowtko, J. McAllister, J. Miller, C. Sotillo, H. Thompson and R.
Weinert (1991) ``The HCRC Map Task Corpus'', {\em Language and Speech\/}
34(4), 351--366.

\xp
Briscoe, T. and J. Carroll (1993) ``Generalized Probabilistic LR Parsing of
Natural Language (Corpora) with Unification-Based Grammars'', {\em
Computational Linguistics\/} 19(1), 25--59.

\xp
Hindle, D. (1989) ``Acquiring Disambiguation Rules from Text'', {\em
Proceedings of the 27th Annual Meeting of the Association for Computational
Linguistics}, 118--125.

\xp
Kay, M. (1989) ``Head-Driven Parsing'', {\em Proceedings of the Workshop on
Parsing Technologies}, 52--62.

\xp
Pereira, F.C.N. and S.M. Shieber (1987) {\em Prolog and Natural Language
Analysis}, CSLI Lecture Notes No.\,10, Stanford University.

\xp
Satta, G. and O. Stock (to appear) ``Bidirectional Context-Free Grammar
Parsing for Natural Language Processing'', {\em Artificial Intelligence}.

\xp
van~Noord, G. (1991) ``Head Corner Parsing for Discontinuous Constituency'',
{\em Proceedings of the 29th Annual Meeting of the Association for
Computational Linguistics}, 114--121.

\end{document}